\documentclass[10pt,conference]{IEEEtran}

\usepackage{cite}

\usepackage{subcaption}
\usepackage{color}
\usepackage{booktabs} 
\usepackage{multirow}
\usepackage[normalem]{ulem}
\usepackage{amsmath,amssymb}

\ifCLASSINFOpdf
   \usepackage[pdftex]{graphicx}
   \graphicspath{{images/}}
   \DeclareGraphicsExtensions{.pdf,.jpeg,.png}
\else
   \usepackage[dvips]{graphicx}
   \graphicspath{{../images/}}
   \DeclareGraphicsExtensions{.eps}
\fi

\usepackage{algorithmic}

\usepackage{array}

\usepackage{url}

\newif\iffinal
\finaltrue

\iffinal
\else
\usepackage[switch]{lineno}
\fi

\begin{document}

%
\title{Improving Mass Detection in Mammography Images: A Study of Weakly Supervised Learning and Class Activation Map Methods}



\iffinal

\author{Vicente Sampaio\inst{1}, Filipe R. Cordeiro \inst{1} }


\author{\IEEEauthorblockN{Vicente Sampaio}
\IEEEauthorblockA{Department of Computing\\
Universidade Federal Rural  de Pernambuco\\ Recife, Brazil\\
Email: vicentegalencar@gmail.com}
\and
\IEEEauthorblockN{Filipe R. Cordeiro}
\IEEEauthorblockA{Visual Computing Lab, Department of Computing  \\
 Universidade Federal Rural  de Pernambuco\\  Recife, Brazil\\
Email: filipe.rolim@ufrpe.br}
}

\else
  \author{Sibgrapi paper ID: 39 \\ }
  \linenumbers
\fi

\maketitle
\let\thefootnote\relax\footnote{\\979-8-3503-3872-0/23/\$31.00
\textcopyright2023 IEEE}

\begin{abstract}
  In recent years, weakly supervised models have aided in mass detection using mammography images, decreasing the need for pixel-level annotations. However, most existing models in the literature rely on Class Activation Maps (CAM) as the activation method, overlooking the potential benefits of exploring other activation techniques. This work presents a study that explores and compares different activation maps in conjunction with state-of-the-art methods for weakly supervised training in mammography images. Specifically, we investigate CAM, GradCAM, GradCAM++, XGradCAM, and LayerCAM methods within the framework of the GMIC model for mass detection in mammography images. The evaluation is conducted on the VinDr-Mammo dataset, utilizing the metrics Accuracy, True Positive Rate (TPR), False Negative Rate (FNR), and False Positive Per Image (FPPI). Results show that using different strategies of activation maps during training and test stages leads to an improvement of the model. With this strategy, we improve the results of the GMIC method, decreasing the FPPI value and increasing TPR.
\end{abstract}

\section{Introduction}




Breast cancer has emerged as the most prevalent cancer affecting women globally. It accounts for a substantial number of cancer-related fatalities, responsible for approximately 15.5\% of all cancer deaths \cite{WHO_BreastCancerPrevention}. Early detection is pivotal in improving treatment outcomes as interventions become more challenging in advanced stages \cite{autier2011advanced}. However, the interpretation of digital mammography images poses significant challenges, even for experienced radiologists, due to various factors, including image quality, radiologist expertise, tissue variations, and lesion characteristics \cite{deshpande2013medical}. To address these challenges and enhance diagnostic accuracy, integrating computer-aided diagnosis (CAD) tools for lesion detection has been recommended to assist radiologists in identifying lesions and defining their boundaries, providing an additional tool to physicians and improving the accuracy of breast cancer diagnosis in mammography images.

Computational approaches leveraging Convolutional Neural Networks (CNNs) have achieved remarkable success in various medical image classification and segmentation tasks~\cite{hesamian2019deep, yadav2019deep}.
 In  cancer diagnosis applications, achieving image interpretability is crucial, and it is accomplished through the localization of key regions in the image that determines the output class assigned by the model~\cite{gmic}, thereby assisting medical professionals in making accurate diagnoses.
 Networks such as U-Net~\cite{ronneberger2015u} and Faster-RCNN~\cite{ren2015faster} have been widely employed for segmentation and detection tasks, with annotations indicating lesion regions and their corresponding classification (benign or malignant). 
 Despite significant advancements in semantic segmentation techniques for medical images, current approaches heavily rely on large training datasets with high-quality annotations to ensure efficient model training~\cite{tran2021tmd}. However, acquiring such datasets poses significant challenges in the medical domain, as lesion annotations necessitate expert knowledge and meticulous annotation of lesion locations in mammograms, making the process labour-intensive and cost-prohibitive~\cite{xie2020instance}.


Given these challenges, the area of weakly supervised learning has been widely studied in recent years \cite{Diba_2017_CVPR, zhang2018adversarial}, exploring strategies to extract information from data with scarce or weak annotations~\cite{ouyang2019weakly}. 
Although there are different levels of weakly supervised learning, in this study, we consider a weakly annotated database as one in which the images have annotation only regarding the image class (normal or with lesion) but that does not have annotation regarding the location or contour of the lesion. This approach facilitates the training of convolutional networks, making the construction and training of models in mammography images more cost-effective and feasible, as it reduces reliance on specialist annotations for lesion localization. Despite advances in weakly supervised learning, this is still an open problem, and new studies aim to improve the results compared to the strongly supervised methods.

Within weakly supervised training methods, the Class Activation Map (CAM)~\cite{cam} technique has been widely employed for detecting lesions in digital mammography images. However, new  activation map-based methods have been proposed and have not been explored in the context of lesion detection within mammography images. This work proposes leveraging weakly supervised learning to study state-of-the-art class activation maps for enhanced lesion detection. Specifically, we compare the effectiveness of CAM, GradCAM~\cite{gradcam}, GradCAM++\cite{gracam++}, XGradCAM\cite{xgradcam}, and LayerCAM~\cite{layercam} methods. Activation maps are evaluated using the state-of-the-art Globally-Aware Multiple Instance Classifier (GMIC)~\cite{gmic} model  and the DRVin-Mamo dataset~\cite{vindrmamo}. The main contributions of this study are outlined as follows:

\begin{itemize}
    \item Exploration of the impact of utilizing different activation map methods for weakly supervised learning in digital mammography images;
    \item Analysis of lesion detection models on the VinDr-Mammo dataset;
    \item Improvement of the GMIC model using different activation maps for training and testing.
\end{itemize}

\section{Related works}

In recent years, several works have been proposed in weakly supervised learning for detecting anomalies in digital mammography images~\cite{gmic}. Among the main models used for weakly supervised detection in mammography images, the CAM method has been extensively employed to identify regions of interest.


Shen et al.~\cite{gmic} propose the GMIC model, which uses a convolutional neural network model incorporating local and global image features. First, this model uses a low-capacity network across the image to identify the most informative regions. Then, a higher-capacity network collects details from the selected regions. Finally, a fusion module aggregates global and local information to make a prediction. The model is trained only with class information of the image, and the regions of interest are obtained using the CAM method.

Liu et al. propose the GLAM method, which builds upon the GMIC model by incorporating refined segmentation using only image-level annotation. The key concept behind GLAM is the selection of informative regions (patches), followed by performing segmentation specifically on these selected regions. Similar to other approaches, GLAM also employs the CAM method for identifying regions of interest.

Liang et al.\cite{liang2020weakly} propose using a CAM activation map to replace old attention models. Additionally, a self-training strategy is utilized, involving the observation of outputs from intermediate layers of the model. Bakalo et al.\cite{bakalo2019classification} adopt a sliding window approach, leveraging a pre-trained VGG network to identify regions of interest for the targeted problem class. While this approach performs well on smaller images, its computational cost escalates significantly when dealing with large databases comprising high-resolution images and deep model training.

Zhu et al. ~\cite{zhu2017deep} tackle region of interest detection by generating a reduced feature map through convolution and max pooling layers. Multiple instance learning (MIL)~\cite{dundar2006multiple} is then employed for image class identification.

Beyond the medical imaging domain, several activation map generation methods have been proposed for weakly supervised learning~\cite{gradcam, xgradcam, gracam++, layercam}. However, the methods applied in digital mammography have been limited to CAM evaluation. Our work analyzes different CAM-based methods proposed in the literature in recent years, showing that the activation map is an important optimization factor in the weakly supervised learning process.

\section{Materials and Methods}

\subsection{Activation Map Methods}


Weakly supervised object detection (WSOD) aims to identify the region containing an object in an image based solely on the image class without pixel-level supervision. Activation map-based methods are commonly employed in WSOD approaches to generate bounding-boxes regions by identifying values above a defined threshold~\cite{infocam}. The resulting region is then resized to match the original image size.


Saliency map methods have been proposed in the literature as an approach to elucidate the relationship between the observed region in the model and the class present in the image~\cite{gmic}. These methods contribute to the interpretability of proposed models and address weakly supervised learning challenges.
Saliency methods based on activation, such as CAM, rely on observing the activation of the final layer of the model to identify the regions responsible for the activation of each class. Activation-based methods have been proposed in medical image classification tasks to assist in the interpretability of the models used~\cite{chestx8, rajpurkar2018deep}. Only the CAM model has been investigated in the context of weakly supervised learning applied to lesion detection in mammography images. However, other approaches have been proposed in the literature in recent years and are analyzed in this work.

\begin{figure*}
    \centering
    \includegraphics[width=1.7\columnwidth]{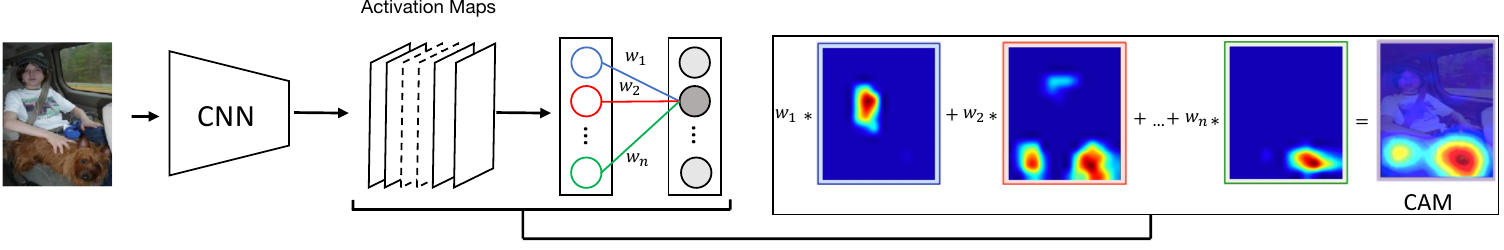}
    \caption{CAM Activation Map. Image adapted from \cite{cam}.}
    \label{fig:cam}
\end{figure*}

Let $f$ be a convolutional neural network with a classifier, and $c$ denotes the class of interest. Given an image $x$ and a convolutional layer $l_i$, where $i$ is the $i$-th convolutional layer of $f$, the class activation map (CAM) of $x$ with respect to $c$ is defined as the linear combination of the activation map $l_i$, as shown below~\cite{poppi2021revisiting}:

\begin{equation}
    CAM_c(x) = ReLU\left (\sum_{k=1}^{N_l}\alpha_kA_k\right ),
\end{equation}

\noindent where $N_l$ represents the number of channels in the convolutional layer $l_i$, $A_k$ is the $k$-th activation channel, and $\alpha_k$ is the weight indicating the importance of the activation channel to class $c$. 
The ReLU activation function is applied to consider only the features that positively influence the target class. The activation map is usually resized to the same size as the input image for CAM-based approaches. Thus, the region of interest can be identified by multiplying the activation map with the input image.
In convolutional networks with a global average pooling layer, the values of $\alpha_k$ correspond to the weights of the final classification layer~\cite{cam}. Figure~\ref{fig:cam} illustrates the process of obtaining the activation map.



The Grad-CAM method~\cite{selvaraju2017grad} determines the coefficient of the activation map by calculating the average gradients across all activation neurons in the map. The Grad-CAM++ method~\cite{gracam++} is a modified version of Grad-CAM that focuses on the positive influences of neurons, considering second-order derivatives. 
The XGradCAM method~\cite{xgradcam} is also based on Grad-CAM but scales the gradients using normalized activations. The LayerCAM method~\cite{layercam} combines activation maps from different layers. According to the authors, the initial layers better capture detailed information about object location, while the deeper layers detect the location of the objects of interest.

\subsection{Training}

\begin{figure}[ht]
    \centering
    \includegraphics[width=1\columnwidth]{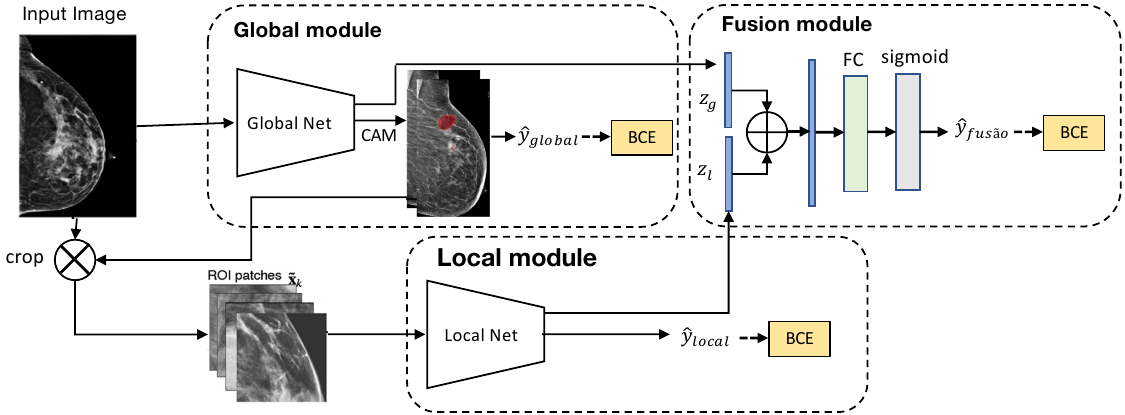}
    \caption{GMIC model. Image adapted from \cite{gmic}.}
    \label{fig:gmic}
\end{figure}

To conduct this study, we employed the GMIC network~\cite{gmic}, a state-of-the-art method for weakly supervised object detection (WSOD) in mammography images. The GMIC model utilizes a global feature extraction module employing CAM to identify regions of interest. These regions are cropped and used as input to a local module. A local feature extraction model extracts the feature vector for each region obtained. Finally, the model is trained by combining the local and global features. Figure \ref{fig:gmic} shows the operation of the GMIC model.



The GMIC loss function is defined by Equation 2 as follows~\ref{eq:gmic}~\cite{gmic}:

\begin{equation}
\begin{aligned}
    L(y,\hat{y}) = \sum_c{ \text{BCE}(y^c, \hat{y}_{local}^c)} + \text{BCE}(y^c, \hat{y}_{global}^c) + \\ \text{BCE}(y_c, \hat{y}_{fusion}^c) + \beta L_{reg}(A^c),
    \label{eq:gmic}
\end{aligned}
\end{equation}

\noindent where BCE represents the binary cross-entropy, $y^c$ denotes the expected output for class $c$, $\hat{y}_{local}^c$, represents the observed output for the local model, $\hat{y}_{global}^c$ corresponds to the observed output for the global model, $\hat{y}_{fusion}^c$ signifies the observed output for the global model after the fusion of local and global features, $\beta$ is a regularization coefficient that employs the activation map $A^c$ according to the $L_{reg}$ function. The regularization function $L_{reg}$ is defined as $L_{reg}=\sum_{i,j}|A_{i,j}^c|$, where $i$ and $j$ represent the rows and columns of the activation map.


\subsection{Experimental Environment}

\begin{figure}[ht!]
\centering
\subfloat[MLO-D]{\includegraphics[width=0.2\columnwidth]{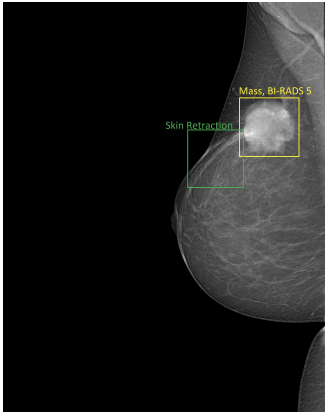} }
\subfloat[CC-D]{\includegraphics[width=0.2\columnwidth]{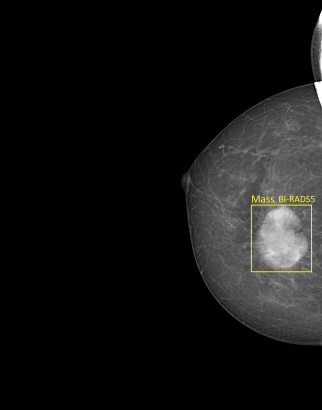} }
\subfloat[CC-E]{\includegraphics[width=0.2\columnwidth]{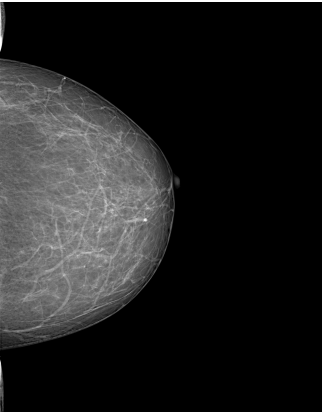} }
\subfloat[MLO-E]{\includegraphics[width=0.2\columnwidth]{Template_SBC/images/base_c.png} }
\caption{Example images from the VinDr-Mamo dataset. CC-D and CC-E labels refer to craniocaudal views of the right and left breast, respectively. MLO-D and MLO-E correspond to mediolateral oblique views of the left and right breast, respectively. The images are sourced from ~\cite{vindrmamo}.}
\label{fig:dataset}
\end{figure}

For model evaluation, we utilized the VinDr-Mammo database~\cite{vindrmamo}, which is publicly available. This database comprises 5000 mammogram exams, each containing four associated images, including two views (mediolateral and craniocaudal) for each breast. The images in the database were acquired using the full-field digital mammography (FFDM) technique. 
The VinDr-Mammo dataset provides information on the anomaly class, such as mass, calcification, asymmetry, and corresponding locations. In our work, we solely used the location information for model validation. During training, only the image class was considered. Specifically, we focused on two classes: "normal" and "mass". The "normal" class signifies that the image does not contain any mass or lesion, while the "mass" class indicates the presence of a lesion potentially associated with a tumour. The training set consisted of 1978 images, and the test set comprised 474. Both sets were balanced in terms of class distribution. Figure~\ref{fig:dataset} presents example images from the dataset.


\subsection{Implementation}

The images from the VinDr-Mamo dataset were resized to a resolution of $2944 \times 1920$. Basic data augmentation techniques were applied to augment the training set, including horizontal flipping, random cropping, and normalization, following the approach used in \cite{gmic}. The training and testing sets were divided based on the dataset's predefined split, selecting the images containing "mass" and "normal" classes. 

To train the GMIC model, a pre-trained model from the NYU Breast Cancer Screening dataset~\cite{wu2019nyu} was utilized, and transfer learning was employed on the VinDr-Mamo dataset. The GMIC model architecture incorporated a ResNet-22~\cite{he2016deep} for the global model and a ResNet-18~\cite{he2016deep} for the local model, as described in~\cite{gmic}. The training process involved 50 epochs, using a $\beta$ value of 3.26 and a batch size 6. The remaining model parameters followed the original values specified by the authors. The code implementation was developed in Python, utilizing the authors' provided source code available on GitHub as the foundation for our work.



For the activation map models  GradCAM, GradCAM++, XGradCAM, and LayerCAM, we based our implementation on the code available at~\cite{jacobgilpytorchcam}. Additionally, we utilized the original code developed in~\cite{gmic} for the CAM model.

\subsection{Avaliation Metrics}

For model evaluation, we employed several metrics to assess the performance of the proposed approach. These metrics included accuracy, Area Under the ROC Curve (AUC), True Positive Rate (TPR), True Negative Rate (TNR), and False Positive per Image (FPPI).


In the classification task, we utilized AUC, TPR, and TNR, commonly used metrics in the literature \cite{ribli2018detecting, shen2019deep}. TPR represents the ratio of correctly classified positive samples, while TNR represents the correct classification of negative samples. The TPR and TNR metrics are defined by equations \ref{eq:tpr} and \ref{eq:tnr}, respectively.

\begin{equation}
    TPR = \frac{\text{TP}}{\text{TP}+\text{FN}},
    \label{eq:tpr}
\end{equation}

\begin{equation}
    TNR = \frac{\text{TN}}{\text{TN}+\text{FP}},
    \label{eq:tnr}
\end{equation}

\noindent where TP, FN, TN, and FP represent the true positives, false negatives, true negatives, and false positives, respectively. In the classification analysis, a true positive occurs when the image class is "mass", and the model correctly predicts it.


We used TPR and FPPI metrics for the detection analysis, commonly used in the literature \cite{jung2018detection, agarwal2020deep}. In the detection model, a predicted location is considered a true positive if the intersection over union (IoU) between the predicted region and the ground truth region is greater than 0.3. The FPPI metric measures the average number of false positive detections per image. Maximizing the TPR rate while minimizing the FPPI rate is the desired outcome.

\section{Results}

Two scenarios were considered to evaluate activation maps using the GMIC model. In the first scenario, the original GMIC model was trained using the CAM method to obtain regions of interest during the training phase. However, different activation map methods were analyzed during the test phase to infer the region's location of interest. The activation map method was changed during the training and test phases in the second scenario. The same training and test sets from the VinDr-Mammo database were used for both scenarios. The obtained metric values after training the original GMIC model are presented in Table.~\ref{tab:acc}:

\begin{table}[ht]
    \centering
    \caption{Results of GMIC model trained using VinDr-Mammo database.}
    \begin{tabular}{c|c|c|c|c}
    \toprule
      Model & Accuracy & AUC & TPR & FNR \\
      \midrule
       GMIC & 80.12 & 87.22 & 71.88 & 88.52 \\
       \bottomrule
    \end{tabular}
    
    \label{tab:acc}
\end{table}

The entire model training was performed using only information from the image class (i.e. normal or with mass). This analysis was done to verify the quality of the model's classification. An accuracy of 80\% indicates that the model can correctly classify most images. This is the first analysis of a weakly supervised model for the VinDr-Mammo dataset.

\begin{figure*}[ht]
    \centering
    \includegraphics[width=1.5\columnwidth]{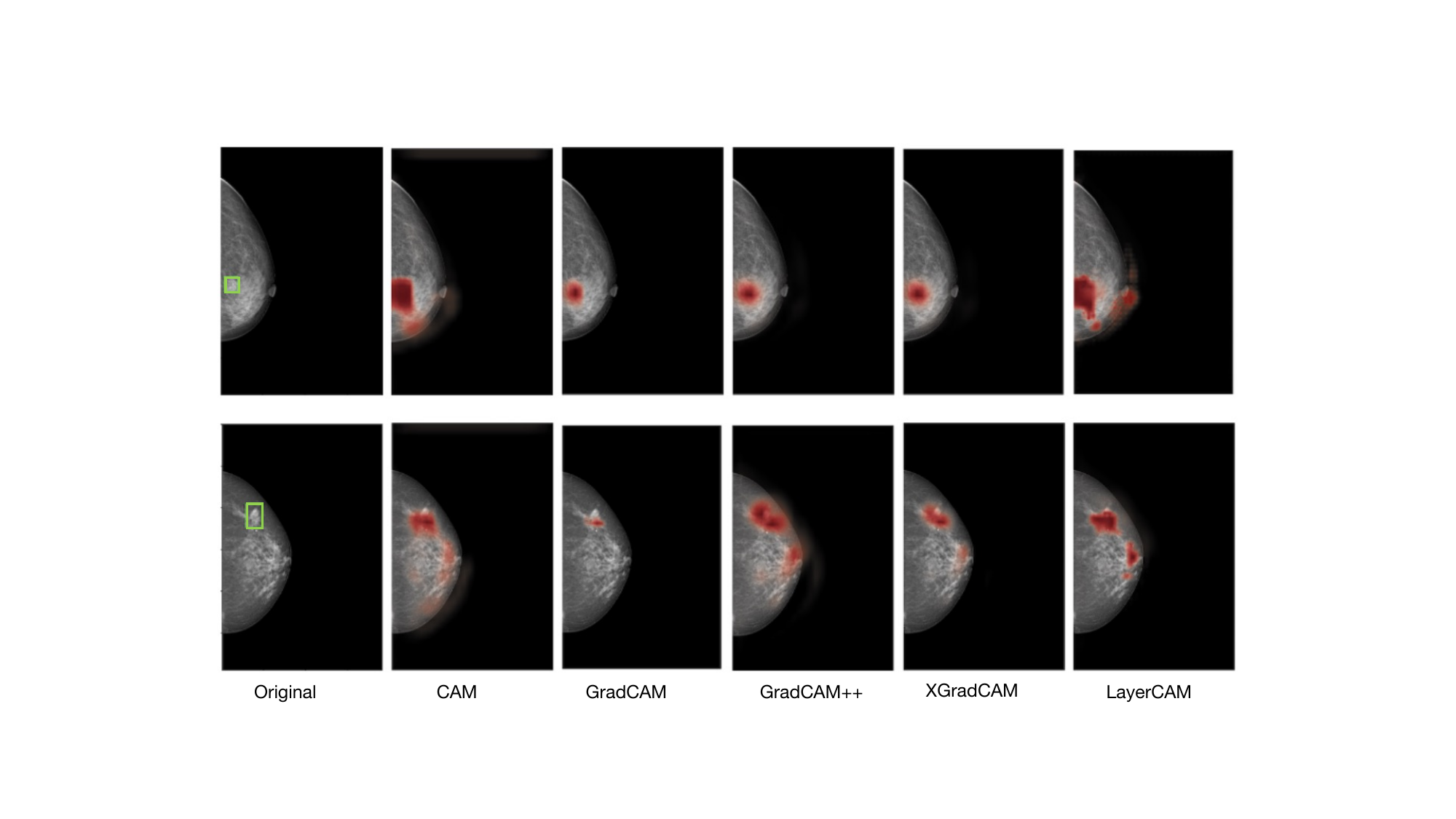} 
    
    \caption{Segmentation of lesion regions using different activation map methods.}
    \label{fig:seg}
\end{figure*}

CAM, GradCAM, GradCAM++, XGradCAM, and LayerCAM models were used to perform lesion region detection. Only the test images containing masses were analyzed to evaluate the detection quality. Figure~\ref{fig:seg} shows the segmentations obtained by each method for two test images. The first column shows the original image, with the ground truth location marked green. Columns 2-6 refer to the CAM, GradCAM, GracCAM++, XGradCAM, and LayerCAM models.

.

Figure~\ref{fig:seg} shows that while all methods can identify the region of interest associated with the lesion, the CAM method tends to generate more false positives, encompassing a larger area of segmented regions. The GradCAM method, on the other hand, produces a much smaller region, occasionally underestimating the size of larger lesions. Although the GradCAM, GradCAM++, and XGradCAM methods yield similar results in Figure~\ref{fig:seg}, a greater distinction between the analyzed methods is observed when considering the entire test set.

Table~\ref{tab:tpr} shows the TPR@FPPI results, which indicate the TPR rate at a specific FPPI value. The highest TPR values obtained were considered for these metrics. Different training and testing scenarios were analysed in Table~\ref{tab:tpr}. The models defined in the rows GMIC (CAM), GMIC (GradCAM++), and GMIC (XGradCAM) represent the results of the GMIC model using the CAM, GradCAM++, and XGradCAM methods during training, respectively. The values in the columns represent the activation map methods used during the inference process in testing. The original GMIC model corresponds to the combination GMIC(CAM)-CAM. Examining the first row, we observe that the original GMIC(CAM)-CAM model achieves the highest TPR rate but with a high FPPI value. Replacing the activation map method can reduce the FPPI rate without significantly decreasing the TPR, as seen when substituting CAM with XGradCAM. Different results in the testing phase are obtained when training GMIC using alternative activation map methods to locate regions of interest. For this analysis, the GMIC (XGradCAM)-GradCAM++ combination yielded the best results, demonstrating a higher TPR and FPPI rate than other methods.

A noteworthy observation from this study is that employing different methods in the training and testing phases can yield improved results compared to using a single model for both stages. We speculate that it is more crucial to have lower detection values with false positives during the training phase, thus enhancing the model's reliability in extracting feature regions. However, a method that generates a larger prediction region during the testing phase leads to a higher TPR value.

In addition, we improved the performance of the GMIC model by replacing the CAM method during training and using the GradCAM++ model during testing. With this, we reduced the FPPI rate from 1.55 to 0.88, increasing the TPR rate.

Additionally, we enhanced the performance of the GMIC model by replacing the CAM method during training with the GradCAM++ model during testing. This substitution reduced the FPPI rate from 1.55 to 0.88, increasing the TPR rate.

\begin{table*}[ht]
    \small
    \centering
    \caption{Results of detection of masses. GMIC model trained with different class activation methods. Results are shown in TPR@FPPI}
    \begin{tabular}{c|c|c|c|c|c}
    \toprule
        Método &  CAM & GradCAM & GradCAM++ & XGradCAM & LayerCAM    \\
        \midrule
        GMIC (CAM) &  0.69@1.55 & 0.60@0.63 & 0.67@1.43 & 0.68@1.05 & 0.68@1.62  \\
        GMIC (GradCAM++) & 0.71@2.89 & 0.13@0.06 & 0.65@1.12 & 0.13@0.06 & 0.67@1.59  \\
        GMIC (XGradCAM) & 0.64@0.75 & 0.60@0.43 & \textbf{0.70@0.88} &0.60@0.42 & 0.72@4.19  \\
        \bottomrule
    \end{tabular}
    
    \label{tab:tpr}
\end{table*}

\section{Conclusion}


This work investigated the impact of different activation map methods on detecting lesions in digital mammography using weakly supervised learning. The results highlighted the significant influence of activation map strategies on the true positive and false positive rates per image, indicating the importance of selecting an appropriate method for lesion detection.


One key finding was that employing different activation maps during the training and testing phases yielded improved inference performance compared to using the same method throughout. By replacing the CAM method with XGradCAM during model training and utilizing GradCAM++ during the testing phase, we reduced the False Positive per Image (FPPI) rate while increasing the model's True Positive Rate (TPR). This modification enhanced the model's ability to detect and localize lesions in mammography images.


For future research, we plan to explore the use of the detections obtained from weakly supervised learning to train the model in a supervised manner. By incorporating this additional information, we aim to refine further and enhance the performance of the detection model. Additionally, we intend to investigate noisy annotation techniques to address incorrect detections during the training process, thereby improving the robustness and reliability of the model's predictions.

\section{Acknowledgment}

 We gratefully acknowledge for financial support of the Brazilian agency Fundação de Amparo à Ciência e Tecnologia do Estado de Pernambuco (FACEPE) with project No. APQ-1046-1.03/21 and  BIC-0067-1.03/22.

\bibliographystyle{IEEEtran}
\bibliography{main}

\end{document}